\documentclass[sn-mathphys,Numbered]{sn-jnl}

\usepackage{graphicx}%
\usepackage{multirow}%
\usepackage{amsmath,amssymb,amsfonts}%
\usepackage{amsthm}%
\usepackage{mathrsfs}%
\usepackage[title]{appendix}%
\usepackage{xcolor}%
\usepackage{textcomp}%
\usepackage{manyfoot}%
\usepackage{booktabs}%
\usepackage{algorithm}%
\usepackage{algorithmicx}%
\usepackage{algpseudocode}%
\usepackage{listings}%

\begin{document}

\title[Article Title]{NewsGPT: ChatGPT Integration for Robot-Reporter}


\author[1]{\fnm{Abdelhadi} \sur{Hireche}}\email{abdelhadi.h@uaeu.ac.ae}
\equalcont{These authors contributed equally to this work.}

\author* [1] {\fnm{Abdelkader Nasreddine} \sur{Belkacem}}\email{belkacem@uaeu.ac.ae}
\equalcont{These authors contributed equally to this work.}

\author[2]{\fnm{Sadia} \sur{Jamil}}\email{sadia.jamil@nottingham.edu.cn}

\author[3,4]{\fnm{Chao} \sur{Chen}}\email{chao\_chen@email.tjut.edu.cn}

\affil*[1]{\orgdiv{College of Information Technology}, \orgname{United Arab Emirates University}, \orgaddress{\state{Abu Dhabi}, \country{United Arab Emirates}}}

\affil[2]{\orgdiv{School of International Communications}, \orgname{University of Nottingham}, \orgaddress{\state{Ningbo}, \country{China}}}

\affil[3]{\orgdiv{Academy of Medical Engineering and Translational Medicine}, \orgname{Tianjin University}, \orgaddress{\state{Tianjin}, \country{China}}}

\affil[4]{\orgdiv{Key Laboratory of Complex System Control Theory and Application}, \orgname{Tianjin University of Technology}, \orgaddress{\state{Tianjin}, \country{China}}}


\abstract{The integration of large language models (LLMs) with social robots has emerged as a promising avenue for enhancing human‒robot interactions at a time when news reports generated by artificial intelligence (AI) are gaining in credibility. This integration is expected to intensify and become a more productive resource for journalism, media, communication, and education. In this paper a novel system is proposed that integrates AI’s generative pretrained transformer (GPT) model with the Pepper robot, with the aim of improving the robot’s natural language understanding and response generation capabilities for enhanced social interactions. By leveraging GPT’s powerful language processing capabilities, this system offers a comprehensive pipeline that incorporates voice input recording, speech-to-text transcription, context analysis, and text-to-speech synthesis action generation. The Pepper robot is enabled to comprehend user queries, generate informative responses with general knowledge, maintain contextually relevant conversations, and act as a more domain-oriented news reporter. It is also linked with a news resource and powered with a Google search capability. To evaluate the performance of the framework, experiments were conducted involving a set of diverse questions. The robot’s responses were assessed on the basis of eight criteria, including relevance, context, and fluency. Despite some identified limitations, this system contributes to the field of journalism and human‒robot interaction by showcasing the potential of integrating LLMs with social robots. The proposed framework opens up opportunities for improving the conversational capabilities of robots, enabling interactions that are smoother, more engaging, and more context aware.}


\maketitle

\section{Introduction}\label{sec1}

The use of large language models with robots has advanced significantly in recent years, enabling robots to engage in more natural and context-aware interactions with humans \cite{wake2023chatgpt}. These language models, such as OpenAI’s GPT (generative pretrained transformer), have demonstrated remarkable capabilities in understanding and generating human-like text \cite{koubaa2023exploring, joublin2023glimpse}. A possible application of this integration is with the Pepper robot, a humanoid robot designed for social interactions and assistance in various domains.

The integration of ChatGPT with the Pepper robot represents a significant advance in human‒robot interaction, bridging the intriguing divide between humans and robots and approaching seamless engagement  \cite{6213238}. Merging the power of an advanced language model with a socially adept robot is leading the way to more intelligent, natural, and context-aware conversations.

One immediate application is in the area of news presentation and delivery. News reporting is experiencing a surge in credibility among audiences due to the incorporation of machine-generated news \cite{haim2017automated}.  This has become an indispensable tool for journalists, serving as a vital resource for their productivity \cite{ford2019newsbots}. With the integration of ChatGPT, many research articles have investigated the creation of robotic software reporters \cite{xu2020xiaomingbot, carlson2015robotic, moravec2020robotic}, While this deal only with software aspects, the Pepper robot has the humanoid appearance and expressive gestures of a person, capable of reporting news and conveying information in a more engaging and natural manner. The advanced language model enables the robot to understand and summarize complex news articles and deliver live news updates quickly, providing the audience with concise and informative reports.
As this technology continues to evolve, it holds the potential to transform the way we interact with robots, making them indispensable companions in various domains, including education, healthcare, and customer service 
 \cite{dwivedi2023so}.

AI-based robot reporters have the potential to replace human reporters for various compelling reasons:

1. Speed and efficiency: ChatGPT-based robots can process vast amounts of information quickly and efficiently. They can rapidly analyze data, generate reports, and disseminate news in real time. This enables faster news delivery, keeping the audience up to date with the latest developments without delays caused by human limitations.

2. Elimination of bias: Human reporters are susceptible to personal biases that can inadvertently affect their reporting. ChatGPT-based robots, being driven by algorithms, are less prone to bias and can provide a more objective representation of the facts. This ensures a fairer and more balanced dissemination of information.

3. Consistency and accuracy: Human journalists may make errors or overlook critical details in their reporting. ChatGPT-based robots, in contrast, can provide consistent and accurate news coverage by relying on their vast knowledge base, fact-checking capabilities, and the ability to cross-reference information from various sources.

4. Availability and scalability: Human reporters have limitations in terms of availability and scalability. They require breaks and sleep, and they cannot cover all news stories simultaneously. ChatGPT-based robots, however, can operate 24/7, tirelessly covering multiple stories simultaneously and ensuring a constant flow of news updates.

5. Language adaptability: ChatGPT-based robots are designed to communicate in multiple languages. This allows them to provide news coverage and reports in different languages, making information more accessible to a global audience. Human reporters, on the other hand, may be limited by their language proficiency.

6. Reduction in costs: Employing human reporters can be expensive, considering salaries, benefits, and training. ChatGPT-based robots offer a cost-effective alternative as they do not require the same level of compensation and can be easily scaled without incurring additional expenses.

7. Safety and risk mitigation: Sending human reporters into dangerous or hazardous environments can pose significant risks to their safety. ChatGPT-based robots can be deployed to such locations, collecting real-time data and reporting without endangering human lives.

ChatGPT-based robot reporters can potentially assist human journalists greatly due to their speed, lack of bias, consistency, language adaptability, and cost effectiveness. While the integration of ChatGPT-based robots in reporting has its advantages, human reporters bring into play unique qualities such as emotional intelligence, intuition, and investigative skills. They also possess a deep understanding of societal nuances and can navigate complex ethical dilemmas more effectively. Striking a balance between the use of technology and the value of human journalism is crucial to ensure comprehensive and responsible news coverage. The ChatGPT-based robotic system proposed here can assist journalists by automating certain tasks such as live news reporting, but it cannot replace the invaluable contributions of human intelligence, intuition, empathy, and creativity that journalists bring to their work. The integration of chatbots and robots may enhance certain aspects of news reporting, but it cannot replace the vital role that human journalists play in society.

This paper presents a novel system that leverages the power of GPT integration with the Pepper robot to enhance social interactions and improve the robot’s ability to engage in meaningful conversations and to report news. The proposed framework explores the potential of combining the language generation capabilities of GPT with the physical embodiment and interactive features of the Pepper robot. By integrating these technologies, the aim is to create a more humanlike and engaging experience for users, narrowing the gap between humans and robots.

\section{Methods}
This section provides a comprehensive explanation of all the technical steps involved in developing the NewsGPT-based robot reporter system. Additionally, an experimental paradigm designed to evaluate the performance of this system is introduced. At the first stage, the user initiates an interaction with the Pepper robot by speaking a few words in the form of either a greeting or a question. Subsequently, the Automatic Speech Detector (ASD) system activates the speech recognition service, aiming to procure the transcription of the user's verbal utterance. This transcribed information is then relayed as a request to the AI agent housed on the cloud infrastructure of the proposed system, which operates using the GPT-3.5 model.

\begin{figure}[ht]
  \centering
    \includegraphics[width=1\textwidth]{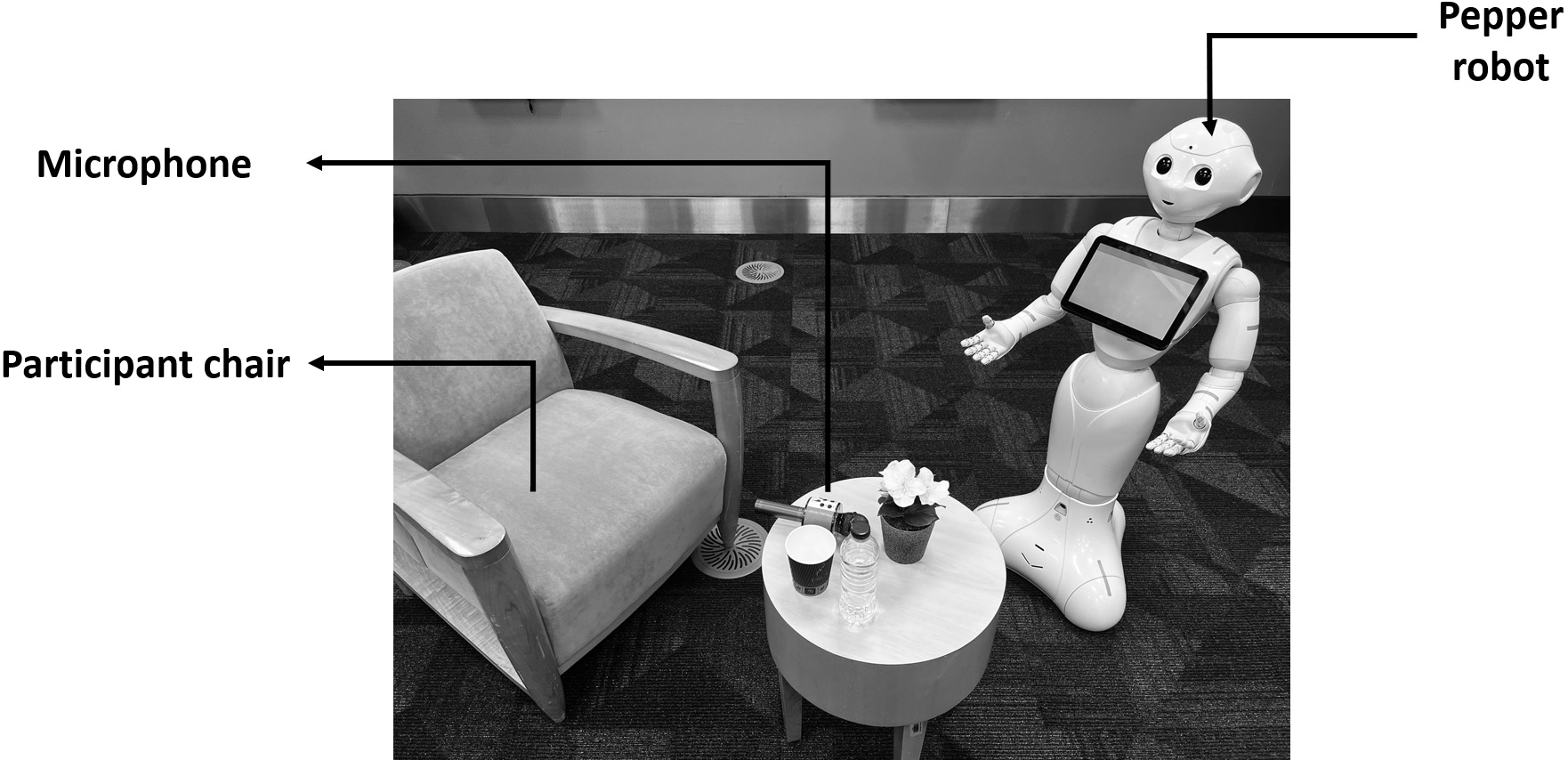}
    \caption{Experiment setup for NewsGPT-based robot-reporter system.}
    \label{fig5}
\end{figure}

\begin{figure}[ht]
\centering
\includegraphics[width=0.7\textwidth]{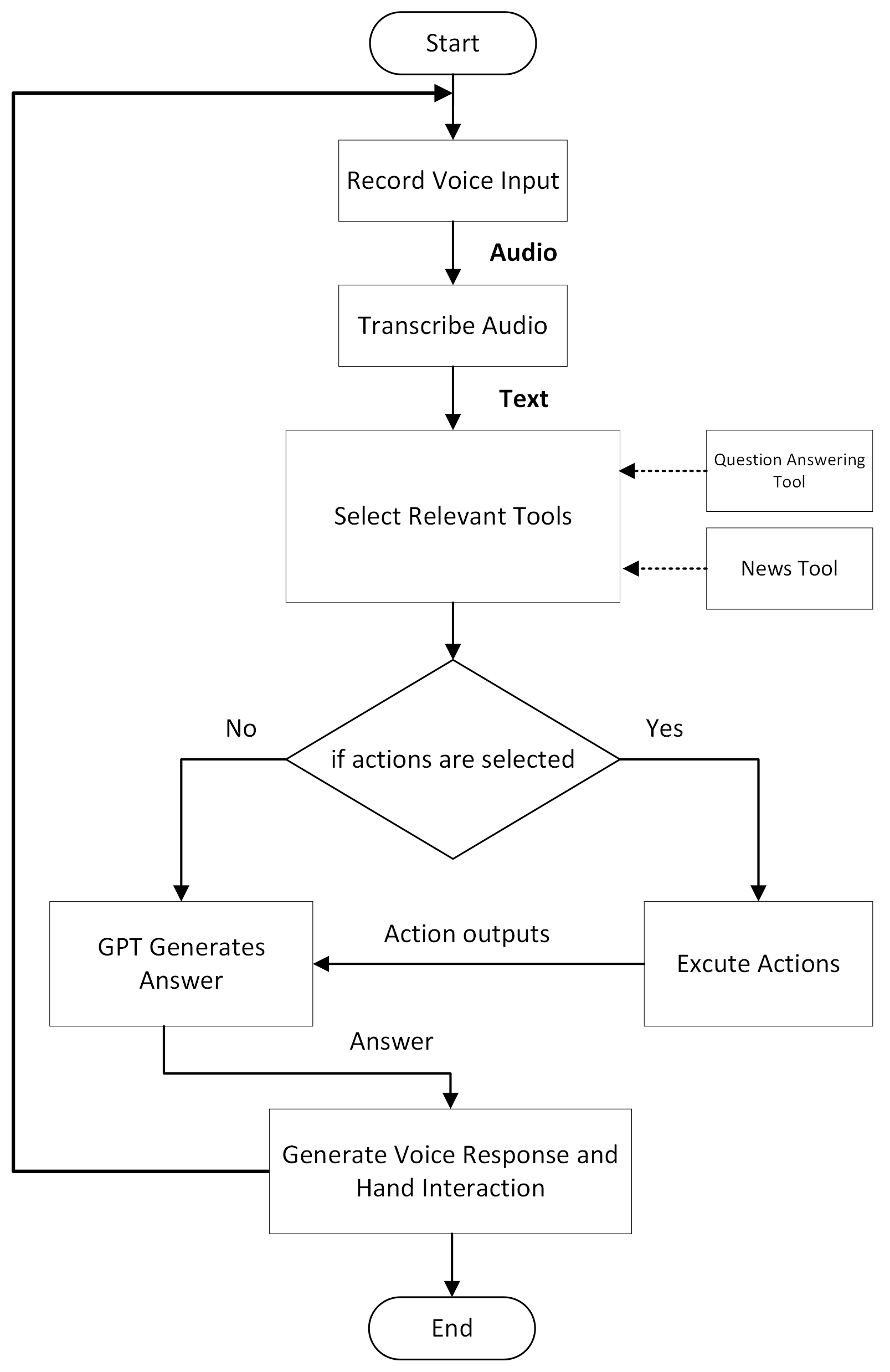}
\caption{Fusion diagram of NewsGPT integration with Pepper robot.}
\label{fig1}
\end{figure}

\begin{figure}[ht]
  \centering
    \includegraphics[width=1.0\textwidth]{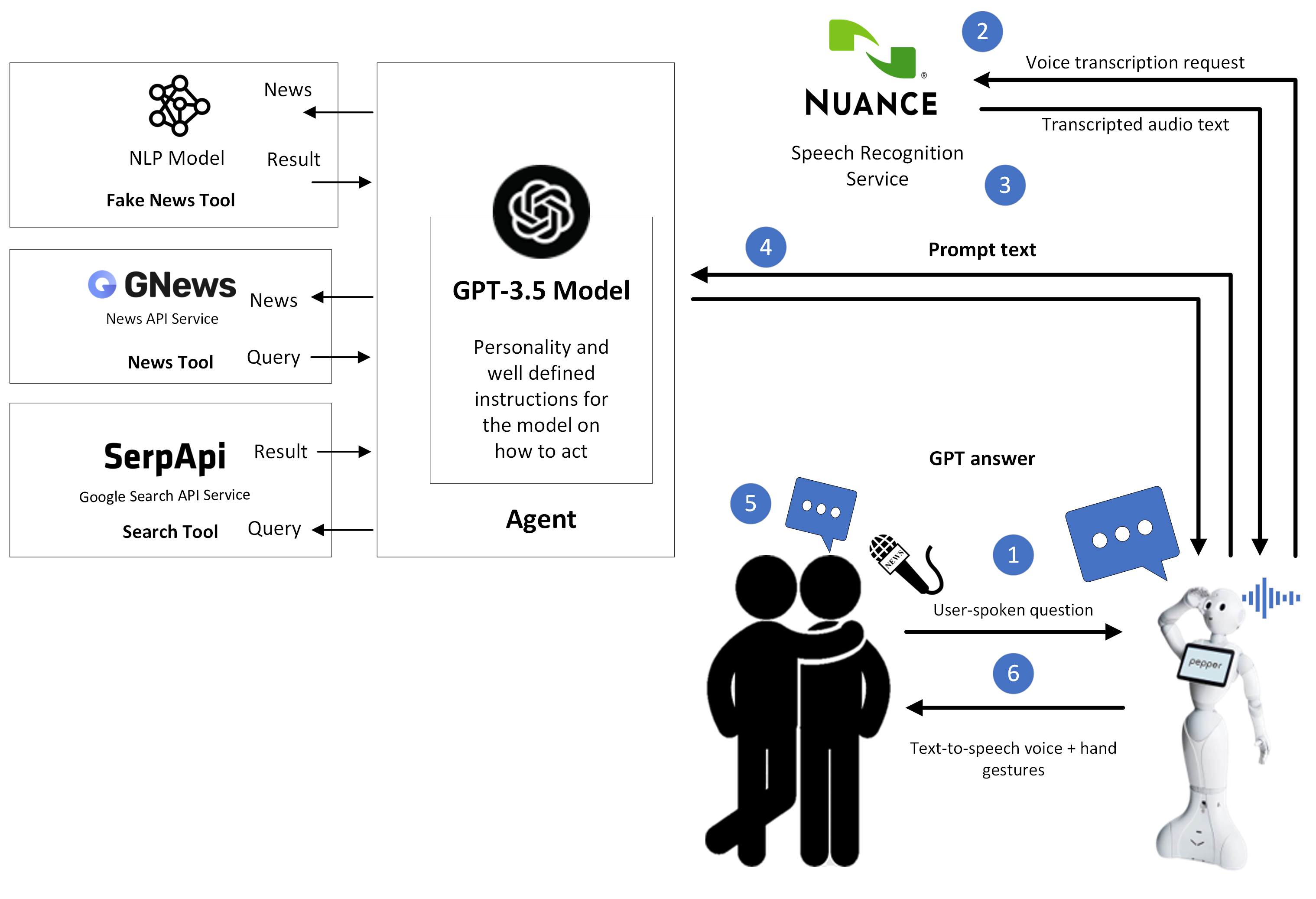}
    \caption{Steps for NewsGPT integration with Pepper at indoor settings: 1) The Pepper robot's blue lights indicate its listening state during the interaction. 2) The robot requests the Nuance Automatic Speech Recognition service to transcribe the user's speech. 3) The Pepper robot sends a request to a cloud-hosted AI agent powered by the GPT 3.5 model. 4) The AI agent used is a ReAct agent, which combines reasoning traces and task-specific actions for interactive and synergistic interactions. 5) The agent incorporates tools such as the Question Answer Tool and the News Tool. 6) The Question Answer Tool retrieves relevant information to respond to user queries. 7) The News Tool uses the GewsAPI to gather recent article headlines based on search terms provided by the model. 8) The retrieved response data from the News Tool is utilized by the language model to construct a news report. 9) The constructed news report is then conveyed back to the user.}
    \label{fig2}
\end{figure}

\begin{figure}[ht]
  \centering
    \includegraphics[width=0.95\textwidth]{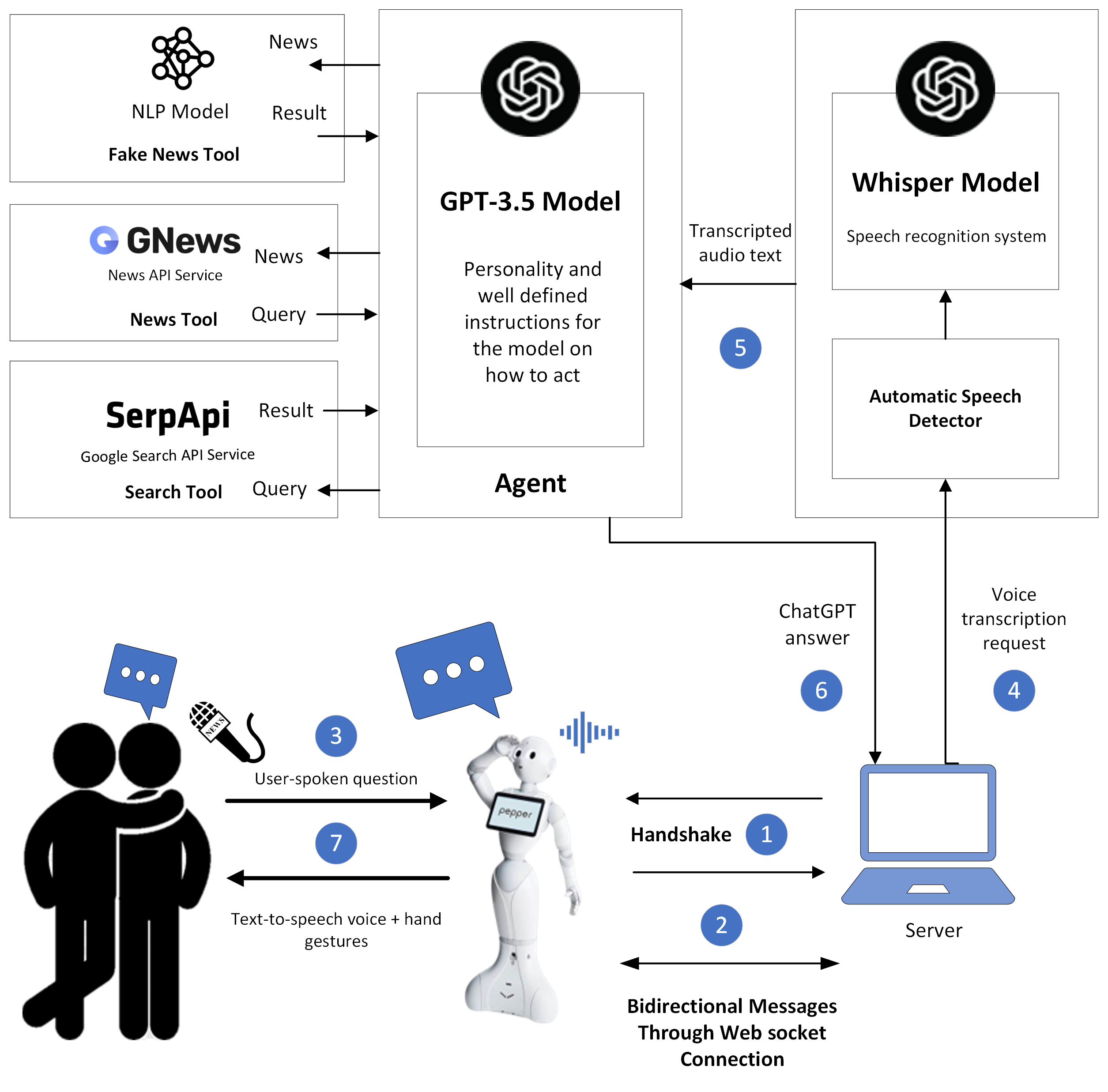}
    \caption{Steps for NewsGPT integration with Pepper at outdoor settings: 1) A laptop-based server manages communication between a cloud AI, ASR, and the Pepper robot while handling chat logic. 2) Pepper, as the client, connects to the server and signals its readiness to listen, which is then conveyed to ASR. 3) ASR records and decodes audio input upon notification from the speech detector. 4) If over 2.5 seconds of silence occurs, the speech detector tells the server to stop recording. The server then retrieves transcribed text from ASR to send to the cloud AI. 5) The AI agent, including GPT-3.5, generates a response. 6) The server relays this response to Pepper, which speaks it to the user.}
    \label{fig3}
\end{figure}

\subsection{AI agent Tools}

AI agents represent an emergent programming paradigm for various AI applications, ranging from question answering and AI management coaching to image generation services and AI companions.

The present study utilized a specific type of agent, known as the ReAct agent, which shares features of the original ReAct agent proposed by Shunyu Yao et al. This ReAct agent enables the generation of both reasoning traces and task-specific actions in an interleaved manner, promoting synergistic interactions between the two. The reasoning traces aid the model in deducing, tracking, and updating action plans, as well as handling exceptional scenarios  \cite{yao2022react}.

To enrich the capabilities of the selected ReAct agent, certain tools were incorporated, such as the Facts Checker Tool, the News Tool and the Search Tool powered by SerpAPI from Google (https://serpapi.com/). These tools are selectively employed by the agent on the basis of the context and received prompts.

the News Tool utilizes the GNewsAPI (https://gnews.io/) to gather recent article headlines related to specific search terms provided by the model tested in this study. The response data retrieved from the News Tool is then utilized by the large language model to construct a news report, which is subsequently conveyed back to the user.

An additional tool is employed to enhance the verification process for checking the authenticity of facts and news. This tool utilizes a deep learning model that is trained on article titles, providing an additional layer of validation.

\subsection{Fake News Detection using deep learning}
To help the robot examine the news articles credibility, we follow an approach to classifying news articles as either fake or real based on their titles. The methodology involves multiple stages, beginning with data preprocessing where two distinct datasets containing fake and real news articles are use together, the dataset is available . each assigned a binary label '0' for fake and '1' for real news. To prepare the dataset for training, each title is truncated or padded to a uniform length, specified as buffer size, and is then transformed into an integer sequence using ASCII values.

The core of our approach lies in the neural network architecture employed for this classification task. The model adopts a sequential framework comprising an input layer to accommodate the preprocessed titles, followed by an embedding layer that maps the ASCII integers into a 64-dimensional space. Subsequently, a 1D convolutional layer with 32 filters and a kernel size of 5 is introduced to capture local sequence patterns. The model also incorporates a flatten layer and a fully connected dense layer with 128 neurons, both utilizing ReLU activation functions. The architecture culminates in an output layer with a single neuron employing a sigmoid activation function to yield probabilities suitable for binary classification.

For the compilation phase, the model is optimized using the adam algorithm and utilizes a binary cross-entropy loss function. Additionally, several evaluation metrics including accuracy, precision, recall, and Area Under the Curve (AUC) are employed to comprehensively assess the model's performance. The model is trained over 10 epochs with a batch size of 64, using a stratified 80-20 train-validation split to maintain class balance.

The table shows the model's performance in classifying news articles as fake or real. With precision, recall, and F1-scores all at 1.00 for both fake and real news, the model excels in making accurate predictions. The balanced dataset, with similar numbers of fake and real articles, adds to the robustness of the evaluation. Overall, the results indicate the model's ability to assess news credibility.

\begin{center}
\begin{table}
\begin{tabular}{cccccc}
\hline
\textbf{Label} & \textbf{Precision} & \textbf{Recall} & \textbf{F1-Score} & \textbf{Support} \\
\hline
Fake & 1.00 & 1.00 & 1.00 & 4696 \\
True & 1.00 & 1.00 & 1.00 & 4284 \\
\hline
\end{tabular}
\caption{Classification Results for Fake News Detection Model.}
\end{table}
\end{center}

\subsubsection{Automatic Speech Recognition}
We utilized two different versions; the first is compatible with the native Nuance ASR that is pre-installed in the Pepper robot, designated for indoor environments where noise is very low. The second version is optimized for noisy conditions, making it suitable for outdoor environments.

The system is composed of two primary elements. Firstly, there's the automatic speech detector, designed to identify the presence of speech by monitoring volume levels. This component initiates the recording process when sounds surpass a specified threshold. Secondly, the system employs the Whisper model, a sophisticated speech recognition service, to accurately interpret and transcribe the captured audio.

The semantic differential (SD) method is a method of measurement that uses subjective ratings of a concept or an object by means of scaling opposite adjectives to study connotative meaning of the concept or object. The technique involves presenting pairs of bipolar, or opposite, adjectives at either end of a series of scales. The difficult part about the semantic differential technique is coming up with words that are truly opposites.

\subsection{Evaluation Procedure}
Table 2 provides a comprehensive evaluation of the ChatGPT-based Pepper conversation experiment, highlighting both its strengths and its areas for improvement. The evaluation of response speed categorized the robot’s performance as either good, average, or poor depending on the time taken to generate a response. Responses completed within 3 seconds were classified as good, while those between 3 and 5 seconds were deemed average. Responses exceeding 5 seconds were classified as poor in terms of speed. Additionally, accuracy was assessed and labeled as either good or bad.

The questions posed to the robot were carefully selected to cover a diverse range of topics and complexities, allowing for a comprehensive assessment of its capabilities. Each answer was evaluated on the basis of the accuracy of the robot’s responses ‒ that is, the extent to which the provided answers aligned with the expected and correct information.

\subsection{Flow of Control}

Evaluating a robot’s conversation involves assessing various aspects of its performance. Key factors to consider are as follows:

\begin{itemize}
\item \textbf{} Relevance: This criterion measures how closely the robot’s responses align with the topic or question asked. It is evaluated by posing general questions, such as inquiring about a country’s capital, and assessing the relevance of the answers provided.

\item \textbf{} Context: This criterion assesses the robot’s ability to retain and recall information over the course of a conversation. It is evaluated by testing whether ChatGPT can remember previously provided information, such as the user’s name, after a certain period of time.

\item \textbf{} Bias: This criterion examines the presence of bias in the robot’s responses. It is evaluated by asking political questions or seeking opinions on controversial figures or topics.

\item \textbf{} Engaging conversation: This criterion measures the robot’s ability to maintain user interest and create engaging conversations. It takes into account factors such as the responsiveness, interactivity, and overall enjoyment of the conversation.

\item \textbf{} Fluency: This criterion evaluates the robot’s performance in terms of speech-to-text functionality, particularly focusing on the accuracy of transcription and the frequency of mistakes made during the conversation.

\item \textbf{} Error resilience: This criterion assesses the robot’s ability to handle and recover from unusual or unexpected inputs. It involves testing the robot’s response to unconventional or nonsensical words or phrases.
\item \textbf{} Domain orientation: This criterion examines the robot’s familiarity with specific domains and its ability to provide information within those domains. It is evaluated by asking the robot about the latest news in a particular country and testing its knowledge of specific past news events.
\item \textbf{} Response time: The robot’s response time is rated over a few different news-related questions and more general knowledge ones.
\item \textbf{} Satisfaction: This criterion measures the level of user satisfaction with the robot’s responses and overall conversation experience. It involves assessing whether the robot’s answers meet the user’s expectations, address their needs, and provide a positive interaction.
\item \textbf{} Creativity: This criterion evaluates the robot’s ability to generate novel and imaginative responses. It assesses whether the robot can offer unique perspectives, suggest creative solutions, or engage in imaginative discussions beyond providing factual information.
\end{itemize}

The numbered steps in Figure 4 signify the following: 

1.	A server, hosted on a laptop, facilitates communication between a cloud-based AI agent, Automatic Speech Recognition (ASR), and the Pepper robot, while also managing chat logic.

2.	The Pepper robot, acting as the client, establishes a connection to the server. Once connected, the robot signals the commencement of listening, which is then relayed to the ASR system.

3.	Upon notification from the automatic speech detector, the ASR initiates the recording and decoding of received auditory input.

4.	If a continuous silence exceeding 2.5 seconds is detected, the automatic speech detector informs the server to cease recording. The server then procures the transcribed text from the ASR to forward to the cloud-hosted AI agent.

5.	Utilizing the GPT 3.5 model, among other tools, the AI agent formulates a response.

6.	This response is relayed by the server to the Pepper robot, which vocalizes the feedback to the user.

Figure 3 demonstrates the functionality of the proposed framework in two specific scenarios. Firstly, when asked about the capital of France, the agent invokes the Search tool, retrieves and analyzes the search results, and subsequently utilizes the obtained information to generate an accurate response. Secondly, when asked about the news in the USA, the agent utilizes the News tool, retrieves relevant news articles, conducts a concise analysis, and presents a brief summary of the news.
These demonstrations showcase the versatility and adaptability of the framework in utilizing different tools depending on the specific enquiry context. The agent effectively leverages the Search tool and the News tool to retrieve and process information, leading to accurate and informative responses. This highlights the capability of the framework in integrating multiple tools seamlessly, enabling the robot to provide users with valuable and relevant information.
The Pepper robot version 1.8a was utilized in the research as the primary hardware platform. The project involved two significant components: the robot Android application, which was developed using the QiSDK API (https://qisdk.softbankrobotics.com), and the cloud agent responsible for processing the language-related tasks.
The Android application served the purpose of providing essential functionalities and facilitating communication with the server. Additionally, an interface server was designed that established a connection between the robot application and the GPT-powered cloud agent.
The cloud agent, which was responsible for processing the language-related tasks, operated with the Python programming language. To facilitate the development of the cloud agent, the LangChain library (https://python.langchain.com) was employed as the core development framework. This library provided essential tools and functionalities to effectively harness the power of the GPT model and enhance the language processing capabilities of the agent.

\subsection{Evaluation Procedure}

Evaluating robot conversation involves assessing various aspects of its performance. Here are some key factors to consider:
\begin{itemize}
\item \textbf{Relevance}: This criterion measures how closely robot responses align with the topic or question asked. It is evaluated by posing general questions, such as inquiring about a country's capital, and assessing the relevance of the answers provided.

\item \textbf{Context}: This criterion assesses robot ability to retain and recall information over the course of a conversation. It is evaluated by testing whether ChatGPT can remember previously provided information, such as the user's name, after a certain period of time.

\item \textbf{Bias}: This criterion examines the presence of bias in the robot responses. It is evaluated by asking political questions or seeking opinions on controversial figures or topics.

\item \textbf{Engaging}: This criterion measures robot's ability to maintain user interest and create engaging conversations. It takes into account factors such as the responsiveness, interactivity, and overall enjoyment of the conversation.

\item \textbf{Fluency}: This criterion evaluates robot's performance in terms of speech-to-text functionality, particularly focusing on the accuracy of transcription and the frequency of mistakes made during the conversation.

\item \textbf{Error resilience}: This criterion assesses robot's ability to handle and recover from unusual or unexpected inputs. It involves testing robot's response to unconventional or nonsensical words or phrases.

\item \textbf{Domain orientation}: This criterion examines robot's familiarity and ability to provide information within specific domains. It is evaluated by asking the robot about the latest news in a particular country and testing its knowledge of specific old news events.

\item \textbf{Response time}: Rate the robot response's time over a few different question related news and more general knowledge ones.

\item \textbf{Satisfaction}: This criterion measures the level of user satisfaction with the robot's responses and overall conversation experience. It involves assessing whether the robot's answers meet the user's expectations, address their needs, and provide a positive interaction.

\item \textbf{Creativity}: This criterion evaluates the robot's ability to generate novel and imaginative responses. It assesses whether the robot can offer unique perspectives, suggest creative solutions, or engage in imaginative discussions beyond providing factual information.
\end{itemize}

Twenty participants were instructed to interact with the robot freely and, after asking a couple of questions, complete a questionnaire (see Appendix 1).

\section{Results and Discussion}

\begin{figure}
\centering
\includegraphics[width=1\textwidth]{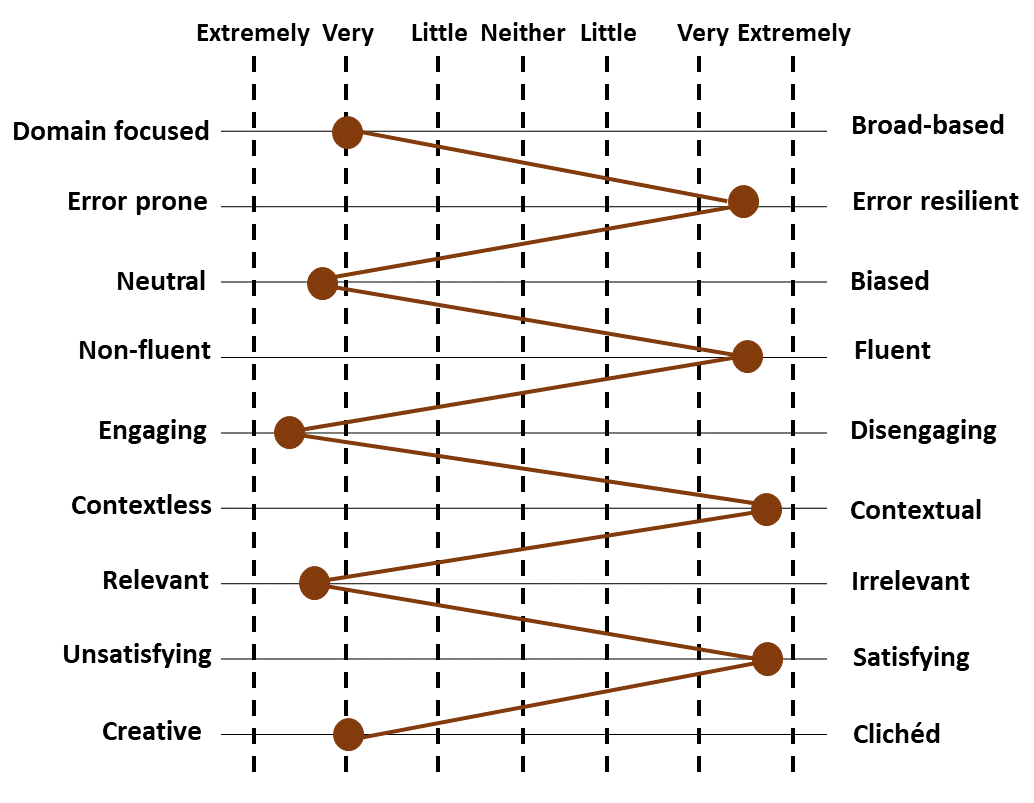}
\caption{Semantic differential scale results to evaluate the user interactions with NewsGPT-based robot reporter.}
\label{fig4}
\end{figure}

To evaluate the proposed NewsGPT-based robot reporter system, an analysis of several metrics was conducted, which is discussed in detail here. Drawing on the experimental results presented in Table 1, a concise assessment in respect of each criterion is provided.

\begin{table}[ht]

\begin{tabular}{|p{0.20\linewidth}|p{0.45\linewidth}|p{0.06\linewidth}|}

\hline
\textbf{\textit{Criteria}} & \textbf{\textit{Evaluation Method}}& \textbf{\textit{Rate}} \\
\hline
Relevance  & Ask a general question, for example about a country’s capital, and see how relevant the answer is. & 3\\
\hline
Context & Tell the robot some specific information, such as a participant’s name, and see if it can remember that information after some time has passed during the conversation. & -2 \\
\hline
Bias & It may not always be possible to eliminate response bias entirely. Therefore, the participant is at liberty to judge how biased the answer was. For example, a political question could be asked, such as “What do you think about President Trump?” & 2 \\
\hline
Engaging & Rate the extent to which the conversation and interactivity with the robot was engaging overall. & 3 \\
\hline
Fluency & Rate the overall experience with speech-to-text functionality and how often mistakes in transcribing were made. & 1 \\
\hline
Error Resilience & Give some bizarre words and see how these are handled. & 3 \\ 
\hline
Domain Orientation & Ask about the latest news in a country, and do a check of specific old news. & 2 \\
\hline
Satisfaction & Evaluate the level of user satisfaction with the robot’s responses and overall conversation experience. & 2 \\
\hline
Creativity & Assess the robot’s ability to generate novel and imaginative responses, offer unique perspectives, and engage in creative discussions. & 2 \\
\hline
\end{tabular}
\caption{ChatGPT-based Pepper conversation experimental results}
\label{tab1}
\end{table}

The ratings of the robot’s performance in terms of the criteria are explained as follows:
\begin{itemize}
\item  \textbf{Relevance}: The relevance of the robot’s responses to general questions, such as about a country’s capital city, was rated 3, indicating a high level of relevance.

\item  \textbf{Context}: The robot’s ability to retain and recall contextual information, such as the user’s name, was rated -2, suggesting that the robot struggled to remember previously provided information over the course of the conversation.

\item  \textbf{Bias}: Bias in the robot’s responses to political questions, such as on President Trump, was rated 2, indicating a moderate level of neutrality in the answers provided.

\item  \textbf{Engaging conversation}: The robot’s ability to maintain user interest and create engaging conversations was rated 3, suggesting that the robot’s responsiveness and interactivity varied and did not consistently create highly engaging conversations.

\item  \textbf{Fluency}: The overall experience with the speech-to-text functionality and the frequency of transcription mistakes were rated 1, indicating that the fluency and accuracy of the speech-to-text functionality were average and had room for improvement.

\item  \textbf{Error Resilience}: The robot’s ability to handle and recover from unusual or unexpected inputs, such as bizarre words, was rated 3, suggesting a reasonable level of error resilience in the robot’s responses.

\item  \textbf{Domain Orientation}: The robot’s familiarity with specific domains, through providing the latest news in a particular country and knowledge of specific old news events, was rated 2, suggesting a moderate level of domain orientation.

\item  \textbf{Response Time}: The average response time across all questions was approximately 7 seconds. Notably, the robot demonstrated prompt responses to general knowledge inquiries, exhibiting a relatively fast response time. Conversely, responses involving complex actions or multi-step processes were characterized by a moderate to poor response time, depending on the number of actions required for generating the response.
\end{itemize}

Table 2 is a record of user questions (UQ) and the corresponding robot answers (RA), along with response speed and accuracy ratings.

\begin{table}[htbp]

\begin{tabular}{|p{0.6\linewidth}|p{0.12\linewidth}|p{0.17\linewidth}|}
\hline
\textbf{\textit{User question (UQ) and robot answer (RA)}} & \textbf{\textit{Response speed (second)}}& \textbf{\textit{Accurateness}} \\
\hline
UQ:	What is the capital of France?

RA:	The capital of France is Paris
 & 7 & +1 (Fact)\\
\hline
UQ:	Who painted the Mona Lisa?

RA:	The mona lisa was painted by leonardo da vinci
 & 10 & +1 \\
\hline
UQ:	What is the boiling point of water? 

RA:	the boiling point of water is 100 degree Celsius at the standard atmospheric pressure & 3 & +1 \\
\hline
UQ:	How many planets are there in our solar system?

RA:	in 2006 the number of planets in the solar system was reduced to 8 where pluto was classified as a dwarf planet 
 & 9 & +1 \\
\hline
UQ:	What is the highest mountain in the world?

RA:	The highest mountain in the world is moount everest standing at 29028 feet above sea level however the tallest mountain from base to top is the tallest mountain from base to summit is Maria in Hawaii. in Huwaii 
 & 9 & +1 \\
\hline
UQ:	What’s the news in USA?

RA:	The latest news in the USA includes Jamie Red and his wife leaving the UK for the USA. 
Bob Aon considering to surprise opponents for Tyson Fury’s next fight.
The death of Ted Kaczynski in prison in a long jumper named M Lewis Shankar finishing third Diamond League.
Also there’s news about a a voice of Wada has been creating waves and R
league in the USA.
 & 10 & 0 (The answer is subjective because the question was not specific. It depends on the date and other GNews selection criteria.
)\\
\hline

UQ:	What’s the news in Egypt?
RA:	Egypt’s health ministry announced 1527 new Coronavirus infections and 92 deaths.
 & 7 & 0 (The robot didn't mention the data of this info.) \\
\hline
UQ:	What’s the latest sport news?

RA:	The latest sports news includes updates on the transfer window, fantasy leagues and various teams dealings with investors and sponsors. It seems there are a lot of changes happening in the sports world. & 6 & -1 (Not enough as an answer.) \\
\hline

\end{tabular}

\caption{*Regarding accuracy, the value +1 means an accurate answer, 0 means a moderately accurate answer, and -1 means an inaccurate answer.}
\label{tab2}
\end{table}

The following conclusions have been drawn from the data in respect of the robot’s response speed, accuracy, and content analysis:
Response speed: The response speed of the robot varied for different questions, ranging from 3 to 10 seconds (s). A response speed of 10 s indicates a relatively slower response, while a response speed of 3 s indicates a faster response. It is important to note that response speed may be affected by various factors, such as system load, network latency, and the complexity of the question.
Accuracy: The accuracy of the robot’s answers is consistently rated as +1, indicating that the answers provided by the robot are considered accurate by the evaluator. This suggests that the model has been fashioned well and is able to provide correct information for a wide range of factual questions.
Content analysis of answers: The robot’s answers seemed to be appropriate and accurate for most of the questions. It correctly identified the capital of France as Paris, the painter of the Mona Lisa as Leonardo da Vinci, the boiling point of water as 100 degrees Celsius, and the current number of planets in the solar system as 8 (with Pluto being classified as a dwarf planet). It also correctly identified Mount Everest as the highest mountain in the world, although there is additional information about the tallest mountain from base to summit being Mauna Kea in Hawaii. The robot’s response to the question about the news in the USA seemed unfocused and confusing. It mentioned various names and events without providing any context or meaningful information. The accuracy rating for this answer is 0, indicating that the evaluator found it to be inaccurate or irrelevant.
In conclusion, the robot generally provided accurate answers to very specific questions, with varying response speeds. However, in the case of broad news-related questions, the robot’s response appeared imprecise and unhelpful. This suggests that the model may not be well equipped to generate meaningful and up-to-date news summaries. 
The source code pertaining to the present research project has been made available for accessing the open source repository on GitHub via the URL: https://github.com/aeh1707/NewsGPT\_Pepper. Additionally, a demonstration video showcasing the project’s functionality is published at: https://shorturl.at/hAIN5

\section{Conclusion}
The integration of OpenAI’s GPT model with the Pepper robot presents exciting possibilities for human‒robot interaction and language capabilities. The NewsGPT system, showcased through the robot reporter, demonstrates the potential of using large language models in journalism. By leveraging GPT’s language processing capabilities, the Pepper robot can deliver news stories in a manner that is context aware and engaging. With the ability to comprehend complex articles and generate informative responses, the robot can provide live news updates and deliver concise reports to the audience. While the NewsGPT-based robot reporter offers advantages like speed, efficiency, and elimination of bias, it is important to acknowledge the unique qualities of human journalists, such as emotional intelligence and investigative skills, to ensure comprehensive and responsible news coverage.
In conclusion, the NewsGPT framework detailed here represents a step forward in human‒robot interaction, bridging the gap between humans and robots in intelligent and interactive conversations. The integration of ChatGPT with the Pepper robot opens up opportunities for more engaging and context-aware interactions in various domains. However, it is crucial to strike a balance between the use of technology and the value of human journalism, as human reporters possess essential qualities such as intuition and understanding of societal nuances. As this technology continues to evolve, ethical considerations must be explored to ensure responsible use for the benefit of society as a whole in domains such as education, healthcare, and customer service.



\begin{appendices}

\section{}\

\textbf{Supplementary Table 1. Questionnaire for robot-reporter performance evaluation.}

\textbf{Question 1: Demographic subquestions}

– Gender

- Age

– Education level

\textbf{Question 2:} Which entity would you prefer to interact with when seeking general information or receiving news updates? \

1. Individual of the same gender \

2. Individual of the opposite gender \

3. Robot \

\textbf{Question 3} how relevant do you find the robot's responses? \

\textbf{Question 4}: How engaged do you feel when interacting with the robot? \

\textbf{Question 5}: How biased do you perceive the robot's responses to be? \

\textbf{Question 6}: How would you rate the robot's fluency and understanding of language? \

\textbf{Question 7}: How aware of the context do you think the robot is during the conversation? \

\textbf{Question 8}: How resilient to errors do you find the robot during the conversation? \

\textbf{Question 9}: How satisfied are you with the robot's news reporting during the conversation? \

\textbf{Question 10}: Do you find the idea of integrating ChatGPT with robot creative for reporting news? \




\end{appendices}


\section*{Declarations}

\begin{itemize}
\item Conflict of Interest: The authors declare that they have no conflict of interest.
\item Dataset on fake news is publicly available: \url{https://www.kaggle.com/c/fake-news/data}
\end{itemize}

\bibliography{sn-bibliography}


\begin{thebibliography}{11}
\ifx \bisbn   \undefined \def \bisbn  #1{ISBN #1}\fi
\ifx \binits  \undefined \def \binits#1{#1}\fi
\ifx \bauthor  \undefined \def \bauthor#1{#1}\fi
\ifx \batitle  \undefined \def \batitle#1{#1}\fi
\ifx \bjtitle  \undefined \def \bjtitle#1{#1}\fi
\ifx \bvolume  \undefined \def \bvolume#1{\textbf{#1}}\fi
\ifx \byear  \undefined \def \byear#1{#1}\fi
\ifx \bissue  \undefined \def \bissue#1{#1}\fi
\ifx \bfpage  \undefined \def \bfpage#1{#1}\fi
\ifx \blpage  \undefined \def \blpage #1{#1}\fi
\ifx \burl  \undefined \def \burl#1{\textsf{#1}}\fi
\ifx \doiurl  \undefined \def \doiurl#1{\url{https://doi.org/#1}}\fi
\ifx \betal  \undefined \def \betal{\textit{et al.}}\fi
\ifx \binstitute  \undefined \def \binstitute#1{#1}\fi
\ifx \binstitutionaled  \undefined \def \binstitutionaled#1{#1}\fi
\ifx \bctitle  \undefined \def \bctitle#1{#1}\fi
\ifx \beditor  \undefined \def \beditor#1{#1}\fi
\ifx \bpublisher  \undefined \def \bpublisher#1{#1}\fi
\ifx \bbtitle  \undefined \def \bbtitle#1{#1}\fi
\ifx \bedition  \undefined \def \bedition#1{#1}\fi
\ifx \bseriesno  \undefined \def \bseriesno#1{#1}\fi
\ifx \blocation  \undefined \def \blocation#1{#1}\fi
\ifx \bsertitle  \undefined \def \bsertitle#1{#1}\fi
\ifx \bsnm \undefined \def \bsnm#1{#1}\fi
\ifx \bsuffix \undefined \def \bsuffix#1{#1}\fi
\ifx \bparticle \undefined \def \bparticle#1{#1}\fi
\ifx \barticle \undefined \def \barticle#1{#1}\fi
\bibcommenthead
\ifx \bconfdate \undefined \def \bconfdate #1{#1}\fi
\ifx \botherref \undefined \def \botherref #1{#1}\fi
\ifx \url \undefined \def \url#1{\textsf{#1}}\fi
\ifx \bchapter \undefined \def \bchapter#1{#1}\fi
\ifx \bbook \undefined \def \bbook#1{#1}\fi
\ifx \bcomment \undefined \def \bcomment#1{#1}\fi
\ifx \oauthor \undefined \def \oauthor#1{#1}\fi
\ifx \citeauthoryear \undefined \def \citeauthoryear#1{#1}\fi
\ifx \endbibitem  \undefined \def \endbibitem {}\fi
\ifx \bconflocation  \undefined \def \bconflocation#1{#1}\fi
\ifx \arxivurl  \undefined \def \arxivurl#1{\textsf{#1}}\fi
\csname PreBibitemsHook\endcsname

\bibitem[\protect\citeauthoryear{Wake et~al.}{2023}]{wake2023chatgpt}
\begin{botherref}
\oauthor{\bsnm{Wake}, \binits{N.}},
\oauthor{\bsnm{Kanehira}, \binits{A.}},
\oauthor{\bsnm{Sasabuchi}, \binits{K.}},
\oauthor{\bsnm{Takamatsu}, \binits{J.}},
\oauthor{\bsnm{Ikeuchi}, \binits{K.}}:
Chatgpt empowered long-step robot control in various environments: A case
  application.
arXiv preprint arXiv:2304.03893
(2023)
\end{botherref}
\endbibitem

\bibitem[\protect\citeauthoryear{Koubaa et~al.}{2023}]{koubaa2023exploring}
\begin{botherref}
\oauthor{\bsnm{Koubaa}, \binits{A.}},
\oauthor{\bsnm{Boulila}, \binits{W.}},
\oauthor{\bsnm{Ghouti}, \binits{L.}},
\oauthor{\bsnm{Alzahem}, \binits{A.}},
\oauthor{\bsnm{Latif}, \binits{S.}}:
Exploring chatgpt capabilities and limitations: A critical review of the nlp
  game changer
(2023)
\end{botherref}
\endbibitem

\bibitem[\protect\citeauthoryear{Joublin et~al.}{2023}]{joublin2023glimpse}
\begin{botherref}
\oauthor{\bsnm{Joublin}, \binits{F.}},
\oauthor{\bsnm{Ceravola}, \binits{A.}},
\oauthor{\bsnm{Deigmoeller}, \binits{J.}},
\oauthor{\bsnm{Gienger}, \binits{M.}},
\oauthor{\bsnm{Franzius}, \binits{M.}},
\oauthor{\bsnm{Eggert}, \binits{J.}}:
A glimpse in chatgpt capabilities and its impact for ai research.
arXiv preprint arXiv:2305.06087
(2023)
\end{botherref}
\endbibitem

\bibitem[\protect\citeauthoryear{Mori et~al.}{2012}]{6213238}
\begin{barticle}
\bauthor{\bsnm{Mori}, \binits{M.}},
\bauthor{\bsnm{MacDorman}, \binits{K.F.}},
\bauthor{\bsnm{Kageki}, \binits{N.}}:
\batitle{The uncanny valley [from the field]}.
\bjtitle{IEEE Robotics \& Automation Magazine}
\bvolume{19}(\bissue{2}),
\bfpage{98}--\blpage{100}
(\byear{2012})
\doiurl{10.1109/MRA.2012.2192811}
\end{barticle}
\endbibitem

\bibitem[\protect\citeauthoryear{Haim and Graefe}{2017}]{haim2017automated}
\begin{barticle}
\bauthor{\bsnm{Haim}, \binits{M.}},
\bauthor{\bsnm{Graefe}, \binits{A.}}:
\batitle{Automated news: Better than expected?}
\bjtitle{Digital journalism}
\bvolume{5}(\bissue{8}),
\bfpage{1044}--\blpage{1059}
(\byear{2017})
\end{barticle}
\endbibitem

\bibitem[\protect\citeauthoryear{Ford and Hutchinson}{2019}]{ford2019newsbots}
\begin{barticle}
\bauthor{\bsnm{Ford}, \binits{H.}},
\bauthor{\bsnm{Hutchinson}, \binits{J.}}:
\batitle{Newsbots that mediate journalist and audience relationships}.
\bjtitle{Digital Journalism}
\bvolume{7}(\bissue{8}),
\bfpage{1013}--\blpage{1031}
(\byear{2019})
\end{barticle}
\endbibitem

\bibitem[\protect\citeauthoryear{Xu et~al.}{2020}]{xu2020xiaomingbot}
\begin{botherref}
\oauthor{\bsnm{Xu}, \binits{R.}},
\oauthor{\bsnm{Cao}, \binits{J.}},
\oauthor{\bsnm{Wang}, \binits{M.}},
\oauthor{\bsnm{Chen}, \binits{J.}},
\oauthor{\bsnm{Zhou}, \binits{H.}},
\oauthor{\bsnm{Zeng}, \binits{Y.}},
\oauthor{\bsnm{Wang}, \binits{Y.}},
\oauthor{\bsnm{Chen}, \binits{L.}},
\oauthor{\bsnm{Yin}, \binits{X.}},
\oauthor{\bsnm{Zhang}, \binits{X.}}, et al.:
Xiaomingbot: A multilingual robot news reporter.
arXiv preprint arXiv:2007.08005
(2020)
\end{botherref}
\endbibitem

\bibitem[\protect\citeauthoryear{Carlson}{2015}]{carlson2015robotic}
\begin{barticle}
\bauthor{\bsnm{Carlson}, \binits{M.}}:
\batitle{The robotic reporter: Automated journalism and the redefinition of
  labor, compositional forms, and journalistic authority}.
\bjtitle{Digital journalism}
\bvolume{3}(\bissue{3}),
\bfpage{416}--\blpage{431}
(\byear{2015})
\end{barticle}
\endbibitem

\bibitem[\protect\citeauthoryear{Moravec et~al.}{2020}]{moravec2020robotic}
\begin{botherref}
\oauthor{\bsnm{Moravec}, \binits{V.}},
\oauthor{\bsnm{Mackov{\'a}}, \binits{V.}},
\oauthor{\bsnm{Sido}, \binits{J.}},
\oauthor{\bsnm{Ek{\v{s}}tein}, \binits{K.}}:
The robotic reporter in the czech news agency: Automated journalism and
  augmentation in the newsroom.
Communication Today
\textbf{11}(1)
(2020)
\end{botherref}
\endbibitem

\bibitem[\protect\citeauthoryear{Dwivedi et~al.}{2023}]{dwivedi2023so}
\begin{barticle}
\bauthor{\bsnm{Dwivedi}, \binits{Y.K.}},
\bauthor{\bsnm{Kshetri}, \binits{N.}},
\bauthor{\bsnm{Hughes}, \binits{L.}},
\bauthor{\bsnm{Slade}, \binits{E.L.}},
\bauthor{\bsnm{Jeyaraj}, \binits{A.}},
\bauthor{\bsnm{Kar}, \binits{A.K.}},
\bauthor{\bsnm{Baabdullah}, \binits{A.M.}},
\bauthor{\bsnm{Koohang}, \binits{A.}},
\bauthor{\bsnm{Raghavan}, \binits{V.}},
\bauthor{\bsnm{Ahuja}, \binits{M.}}, \betal:
\batitle{“so what if chatgpt wrote it?” multidisciplinary perspectives on
  opportunities, challenges and implications of generative conversational ai
  for research, practice and policy}.
\bjtitle{International Journal of Information Management}
\bvolume{71},
\bfpage{102642}
(\byear{2023})
\end{barticle}
\endbibitem

\bibitem[\protect\citeauthoryear{Yao et~al.}{2022}]{yao2022react}
\begin{botherref}
\oauthor{\bsnm{Yao}, \binits{S.}},
\oauthor{\bsnm{Zhao}, \binits{J.}},
\oauthor{\bsnm{Yu}, \binits{D.}},
\oauthor{\bsnm{Du}, \binits{N.}},
\oauthor{\bsnm{Shafran}, \binits{I.}},
\oauthor{\bsnm{Narasimhan}, \binits{K.}},
\oauthor{\bsnm{Cao}, \binits{Y.}}:
React: Synergizing reasoning and acting in language models.
arXiv preprint arXiv:2210.03629
(2022)
\end{botherref}
\endbibitem

\end{thebibliography}

\end{document}